# A Quantized VAE-MLP Botnet Detection Model: A Systematic Evaluation of Quantization-Aware Training and Post-Training Quantization Strategies

Hassan Wasswa
*School of Systems and Computing*
*University of New South Wales*
Canberra, Australia
h.wasswa@unsw.edu.au

Hussein Abbass
*School of Systems and Computing*
*University of New South Wales*
Canberra, Australia
h.abbass@unsw.edu.au

Timothy Lynar
*School of Systems and Computing*
*University of New South Wales*
Canberra, Australia
t.lynar@unsw.edu.au

*Abstract*—In an effort to counter the increasing IoT botnet-based attacks, state-of-the-art deep learning methods have been proposed and have achieved impressive detection accuracy. However, their computational intensity restricts deployment on resource-constrained IoT devices, creating a critical need for lightweight detection models. A common solution to this challenge is model compression via quantization. This study proposes a VAE-MLP model framework where an MLP-based classifier is trained on 8-dimensional latent vectors derived from the high-dimensional train data using the encoder component of a pretrained variational auto-encoder (VAE). Two widely used quantization strategies—Quantization-Aware Training (QAT) and Post-Training Quantization (PTQ)—are then systematically evaluated in terms of their impact on detection performance, storage efficiency, and inference latency using two benchmark IoT botnet datasets—N-BaIoT and CICIoT2022. The results revealed that, with respect to detection accuracy, the QAT strategy experienced a more noticeable decline, whereas PTQ incurred only a marginal reduction compared to the original unquantized model. Furthermore, PTQ yielded a 6× speedup and 21× reduction in size, while QAT achieved a 3× speedup and 24× compression, demonstrating the practicality of quantization for device-level IoT botnet detection.

## I. INTRODUCTION

The Internet of Things (IoT) has expanded rapidly across various domains, fostering progress in automation and connectivity. However, this rapid evolution has also led to increased exposure to cybersecurity threats. Among these, IoT botnet-based attacks represent a critical challenge, impacting both individual users and organizational infrastructures. To counter such threats, a wide range of learning-based detection approaches have been developed, spanning from relatively simple artificial neural networks such as MLPs to more advanced deep architectures, including CNN-LSTMs [1]–[3], ResNets [4], [5], CNN-GRUs [6]–[8], graph neural networks [9]–[12], and transformer-based models with multi-head attention [13]–[16] and many more. These methods have demonstrated near-perfect accuracy in identifying malicious activities.

However, despite their success, deep neural networks remain computationally demanding, which poses obstacles for their deployment on IoT edge devices that operate under stringent resource constraints, including limited power, memory, and processing capabilities. Such models typically require significant computational resources, large memory capacity, and considerable storage for intermediate processing. With the continuous evolution of cyber threats targeting IoT infrastructure, there is a growing need for inference techniques capable of maintaining strong detection performance while significantly lowering resource consumption to enable efficient on-device execution.

The computational demands of deep learning models are largely determined by factors such as input dimensionality, the number of trainable parameters (weights), and the size of activation maps [17]. To address the challenge of high-dimensional inputs, methods such as feature selection [18]–[22], and deep learning-based dimensionality reduction [23]–[25] have been proposed. On the other hand, to alleviate the complexity associated with large parameter sets and extensive activation volumes, quantization methods—particularly Quantization-Aware Training (QAT) and Post-Training Quantization (PTQ)—have been widely explored and applied [26]–[30].

This paper focuses on the quantization strategies which are designed to enhance computational efficiency through the use of low-bit weights and activations [31], [32], while preserving acceptable accuracy levels. Specifically, we systematically evaluate the effectiveness of the two quantization strategies—QAT and PTQ—in terms of their impact on the attack detection performance, disk storage requirement, and detection latency of a VAE-MLP model. The VAE-MLP model involves training a VAE, using its encoder model to project the high-dimensional train features to a latent space dimension and followed by training and evaluating an MLP model on the low dimensional vectors.

Results indicate that PTQ yields a 6× speedup and 21× reduction in size, while QAT achieves a 3× speedup and 24× compression, demonstrating the practicality of quantization for device-level IoT botnet detection.

The remainder of this paper is organized as follows. Section II reviews the related literature, while Section III provides a detailed description of the proposed methodology. Experimental results are presented in Section IV and discussed in Section V. Finally, Section VI summarizes the key findings and concludes the paper.

## II. RELATED WORK

### A. Model Quantization

Quantization refers to reducing the numerical precision of weights and intermediate computations, commonly by converting 32-bit floating-point (FP32) representations to lower-precision formats such as FP16 or INT8. This reduction is particularly relevant for deployment on resource-constrained platforms, such as IoT edge devices, since it lowers the number of floating-point operations (FLOPs), thereby accelerating inference [33]–[35].

According to studies [36], [37], INT8 quantization, in particular, has gained widespread attention since it allows models to achieve accuracy comparable to FP32 representations, while reducing model size by a factor of four and enabling faster inference due to the broad hardware support for 8-bit operations. Given the strict limitations of memory, computation, and energy in these devices, quantization provides key benefits including model compression, faster inference, and reduced power usage, all of which are vital for real-time on-device intelligence in smart sensors, embedded systems, and edge computing. The following subsections review prior studies on: (1) quantization-aware training (QAT) and (2) post-training quantization (PTQ).

### B. Quantization-aware training

Quantization-aware training (QAT) integrates the quantization process directly into model training, enabling the network to adapt to quantization-induced noise [31], [32]. QAT has been widely employed in efforts to develop lightweight models suitable for IoT and WSN devices. For example, [38] combined QAT with neural architecture search to design a neural network that minimizes computational cost, particularly regarding hardware usage and inference latency. In [39], a quantization-aware framework was proposed to account for errors introduced by digital- to-analog (DAC) and analog-to-digital converters (ADCs) in photonic neural networks. Additionally, [40] applied QAT alongside approximation-aware training to optimize a GNN model for a lower operations-to-parameter ratio, achieving faster inference. One drawback of QAT is that despite these advantages, it generally increases computational demands during the training phase.

### C. Post-training quantization

Post-training quantization (PTQ) involves converting a pretrained model into a lower-precision format without retraining, offering a straightforward and rapid approach. However, models sensitive to precision loss may experience accuracy degradation. Numerous studies have investigated PTQ [41]–[45]. For instance, [41] proposed a PTQ method for vision transformers to reduce memory usage and computational load, formulating quantization as an optimization problem to identify optimal low-bit quantization intervals for weights and inputs. Similarly, [43] applied PTQ to reduce diffusion models to 8-bit precision and introduced two quantized autoencoders—FP16 (QAE-float16) and uint8 (QAE-uint8)—combining pruning, clustering, and quantization for efficient DDoS detection on IoT edge devices.

Study [44] conducted a detailed investigation of post-training quantization techniques for CNNs applied to classification and object detection tasks. A layer-wise quantization strategy that assigns each layer the smallest bit width without incurring accuracy loss was applied, thereby enabling variable-bit compression across the network.

Study [42] proposed a Quantized Deep Neural Network (QDNN) for intrusion detection in IoT environments by integrating CNN with LSTM networks. Their approach employed post-training float16 quantization to reduce computational overhead and memory usage, achieving more than a twofold reduction in model size while preserving detection accuracy. By leveraging the CNN-LSTM architecture's capacity to capture both spatial and temporal features, the model demonstrated enhanced robustness and effectiveness in identifying complex attack patterns.

## III. METHODOLOGY

This study systematically evaluates two quantization strategies: Quantization-Aware Training (QAT) and Post-Training Quantization (PTQ). In both strategies, quantization is applied to compress model weights and activations from FP32 to INT8, as INT8 quantized models typically retain accuracy comparable to their FP32 counterparts [36], [37], while substantially reducing model size and improving computational efficiency.

In the first phase, a VAE-MLP model is proposed in which a variational autoencoder (VAE) with a latent space dimension of 8 is trained, and its encoder is employed to transform high-dimensional training data into low-dimensional latent vector representations. An MLP-based classifier is subsequently trained on these latent vectors for traffic classification. During inference, the test dataset is passed through the encoder to obtain low-dimensional representations, which are then fed into the trained MLP classifier. The model is initially evaluated in terms of detection performance, disk storage requirements, and inference latency prior to quantization.

In the second phase, QAT and PTQ are separately applied, followed by a comparative assessment of detection performance and computational efficiency—specifically disk storage requirements and inference latency—across the original unquantized model, the QAT-optimized model, and the PTQ-optimized model.

**QAT-based strategy**: Quantization-aware training is applied to both the VAE and the classifier. In integer-based QAT, a tensor $v$ is quantized during the forward pass using the expression: $v_q = \min(q_{max}, \max(q_{min}, \lfloor v/s + z \rfloor))$ where $q_{min}$ and $q_{max}$ define the representable range based on bit width and sign configuration [17], [40]. The scale factor, $s$, adjusts tensor values to the quantization range, while the zero-point $z$ maps the real number zero to a quantized integer. Both $s$ and $z$ are scalar parameters optimized during training.

**PTQ-based strategy**: Since inference involves only the encoder and the MLP classifier, post-training quantization is applied to the models trained in phase one. The resulting quantized models are converted into TensorFlow Lite format using tf.lite.TFLiteConverter, enabling efficient deployment on resource-constrained IoT edge de- vices for early botnet detection.

## A. Datasets

The proposed method was assessed using two widely studied public IoT traffic datasets: N-BaIoT [46] and CICIoT2022 [47].

*1) N-BaIoT Dataset:* To overcome the absence of publicly available botnet traffic from real IoT environments, [46] introduced the N-BaIoT dataset. It comprises 115 statistical features (23 features extracted over five time windows) derived from NetFlow packets in a testbed network of nine commercial IoT devices. The dataset contains traffic generated by two major IoT botnet malware families, Mirai and BashLite. Using port mirroring on a physical switch within a realistic network configuration, behavioral data was recorded during the execution of the attacks.

*2) CICIoT2022 Dataset:* In contrast, the CICIoT2022 dataset [47] extends beyond the nine devices of N-BaIoT, encompassing 60 devices across domains such as home automation, IP cameras, and audio systems, utilizing communication protocols including Zigbee, Z-Wave, and WiFi. This diversity provides a broader perspective on IoT traffic behaviors. The dataset was collected over a 30-day period and incorporates a variety of simulated attacks. Feature extraction from the published .pcap files was performed using the updated CICFlowMeter[1], yielding 84 features and over 3.2 million NetFlow instances.

## IV. RESULTS

The experimental results were analyzed with respect to attack detection performance, inference time per instance, and disk storage requirements, providing a comprehensive evaluation of each approach.

[1]https://github.com/GintsEngelen/CICFlowMeter

*1) Detection performance analysis:* The experimental results are summarized in Table I. Performance analysis shows that the QAT technique leads to greater degradation in detection performance across all evaluation metrics, compared to the PTQ strategy, for both datasets. The PTQ approach exhibits a negligible impact on detection performance when compared to the results of the original unquantized model. A visual comparison of the performance outcomes is provided in Figures 1 and 2, corresponding to evaluations conducted on the N-BaIoT and CICIoT2022 datasets, respectively. A similar pattern was reported in study [42] where PTQ was deployed for FP32 to FP16 quantization with no effect on detection performance across accuracy, precision, recall, and F1-Score when the model was evaluated using RT-IOT and NSL-KDD datasets.

TABLE I
DETECTION PERFORMANCE ANALYSIS IN TERMS OF ACCURACY, PRECISION, RECALL AND F1-SCORE FROM THE UNQUANTIZED, QAT, AND PTQ-BASED VAE-DNN MODELS

| Dataset | Quantization Method | Acc | Pr | Rec | F1-score |
|---|---|---|---|---|---|
| N-BaIoT | Unquantized | 0.9980 | 0.9934 | 0.9949 | 0.9941 |
|  | QAT | 0.9683 | 0.9720 | 0.9733 | 0.9717 |
|  | PTQ | 0.9971 | 0.9948 | 0.9947 | 0.9948 |
| CICIoT 2022 | Unquantized | 0.9853 | 0.9889 | 0.8786 | 0.9304 |
|  | QAT | 0.9847 | 0.9885 | 0.8662 | 0.9233 |
|  | PTQ | 0.9853 | 0.9887 | 0.8774 | 0.9297 |

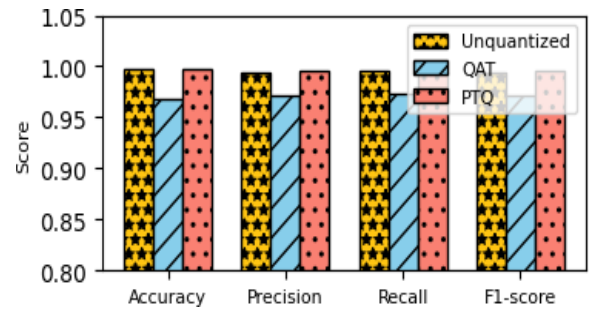

Fig. 1. Detection performance: Unquantized model vs QAT vs PTQ-based models on N-BaIoT dataset across accuracy, precision, recall and F1-score

*2) Storage requirements analysis:* An analysis of the storage requirements revealed that quantization significantly reduces computational costs in terms of disk storage usage. As shown in Table II, the unquantized model required at least 0.431 MB of disk space when trained on the N-BaIoT dataset, and at least 0.430 MB when trained on the CICIoT2022 dataset. In contrast, the QAT- based models required only 0.018 MB and 0.017 MB, while the PTQ-based models required just 0.021 MB for models trained on the N-BaIoT and CICIoT2022 datasets, respectively. The storage analysis is visualized in Figures 3 and 4 for the N-BaIoT and CICIoT2022 dataset respectively.

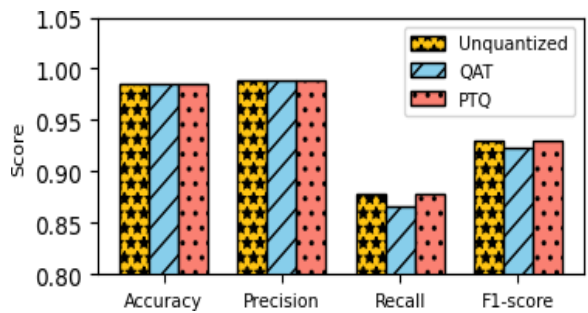

Fig. 2. Detection performance: Unquantized model vs QAT vs PTQ-based models on CICIoT2022 dataset across accuracy, precision, recall and F1-score

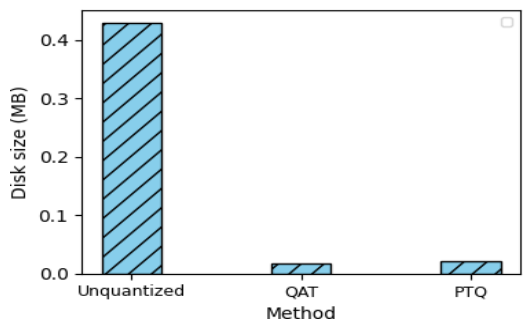

Fig. 4. Disk storage analysis: Models trained ana evaluated on the CICIoT2022 dataset

TABLE II
COST IN TERMS OF STORAGE FOR QUANTIZED AND UNQUANTIZED MODELS

| Dataset | Quantization method | Storage size (MB) |
|---|---|---|
| N-BaIoT | Unquantized | 0.431 |
| | QAT | 0.018 |
| | PTQ | 0.021 |
| CICIoT2022 | Unquantized | 0.430 |
| | QAT | 0.017 |
| | PTQ | 0.021 |

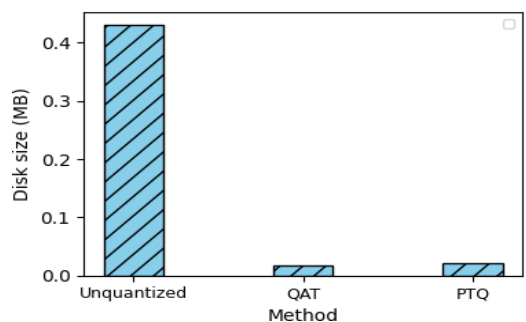

Fig. 3. Disk storage analysis: Models trained ana evaluated on the N-BaIoT dataset

*3) Inference time analysis:* An analysis of the inference time revealed that quantization significantly reduces computational costs in terms of execution latency. As shown in Table III, the unquantized model required at least
$2.388 \times 10^{-5}$ s to perform inference per traffic instance when trained on the N-BaIoT dataset, and $2.382 \times 10^{-5}$ s when trained on the CICIoT2022 dataset. In contrast, the QAT-based models required only $1.016 \times 10^{-5}$ s and $1.011 \times 10^{-5}$ s, while the PTQ-based models achieved even lower inference times of $4.051 \times 10^{-6}$ s and $4.021 \times 10^{-6}$ s for the N-BaIoT and CICIoT2022 datasets, respectively. The inference time comparison is further illustrated in Figures 5 and 6, corresponding to the N-BaIoT and CICIoT2022 datasets, respectively.

TABLE III
COST IN TERMS OF DETECTION LATENCY FOR QUANTIZED AND UNQUANTIZED MODELS

| Dataset | Quantization method | Inference time (s) |
|---|---|---|
| N-BaIoT | Unquantized | $2.388 \times 10^{-5}$ |
| | QAT | $1.016 \times 10^{-5}$ |
| | PTQ | $4.053 \times 10^{-6}$ |
| CICIoT2022 | Unquantized | $2.382 \times 10^{-5}$ |
| | QAT | $1.011 \times 10^{-5}$ |
| | PTQ | $4.021 \times 10^{-6}$ |

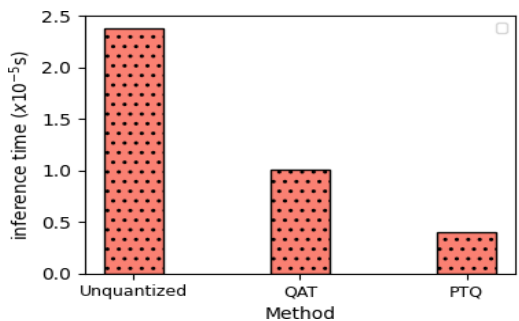

Fig. 5. Inference time per instance: Models trained and evaluated on the N-BaIoT dataset

Given the real-time constraints and limited processing capabilities of IoT devices, this considerable reduction in storage requirements and inference time is critically important for the efficient and practical deployment of AI models in such environments.

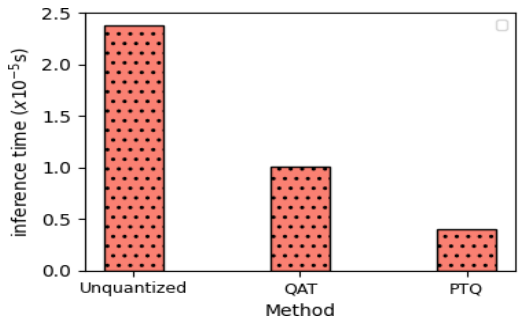

Fig. 6. Inference time per instance: Models trained ana evaluated on the CICIoT2022 dataset

## V. Discussion

Both quantized models exhibit a reduction in classification performance, with the QAT-based approach demonstrating greater performance degradation compared to the PTQ-based method across all evaluation metrics for both datasets. This discrepancy in performance may be partially attributed to the quantization noise introduced during the training phase in QAT, which can affect the quality of the feature representations and subsequently impair the model's learning process. In contrast, the minimal reduction in detection performance observed with the PTQ-based approach suggests that the quantization noise introduced during post-training quantization has a negligible effect on the structure of the learned weight matrices for the various input features.

Moreover, regarding model storage size, the QAT-based model resulted in a smaller file size than the PTQ-based model. This indicates that QAT likely involved more aggressive compression, potentially leading to greater loss of critical classification information—another factor that may explain its comparatively higher performance degradation.

## VI. Conclusion

Although deep learning models deliver strong performance in IoT attack detection, their high computational demands hinder deployment on resource-constrained devices, underscoring the importance of lightweight detection approaches. Reducing computational overhead not only accelerates training and inference but also lowers energy consumption, which is closely tied to runtime efficiency. Common strategies to address these challenges include dimensionality reduction and model compression through quantization. Focusing on the latter, this study systematically evaluates two prominent quantization strategies—Quantization-Aware Training (QAT) and Post-Training Quantization (PTQ)—with respect to detection accuracy, storage efficiency, and inference latency using two benchmark IoT botnet datasets, N-BaIoT and CICIoT2022. The evaluation employs a VAE-MLP architecture, where the encoder component of a pretrained variational autoencoder (VAE) nodes high-dimensional data into 8-dimensional latent vectors that are subsequently classified using an MLP model. Experimental results show that PTQ achieves a 6× speedup with a 21× reduction in model size, while QAT provides a 3× speedup and 24× compression, highlighting the effectiveness of quantization in enabling practical, device-level IoT botnet detection.